\pdfoutput=1

\documentclass[11pt]{article}
\usepackage{authblk}
\usepackage[]{acl}

\usepackage{times}
\usepackage{latexsym}
\usepackage{arabtex}
\newcommand{\hide}[1]{}
\usepackage{pdfpages}
\usepackage{subfig}
\usepackage{makecell}

\usepackage[T1]{fontenc}

\usepackage[utf8]{inputenc}

\usepackage{microtype}
\usepackage{array}
\newcolumntype{L}{>{\arraybackslash}m{12cm}}
\usepackage{multirow}
\usepackage{amssymb}
\usepackage{physics}

\makeatletter
\newcommand\blfootnote[1]{%
  \begingroup
  \renewcommand{\@makefntext}[1]{\noindent\makebox[1.8em][r]#1}
  \renewcommand\thefootnote{}\footnote{#1}%
  \addtocounter{footnote}{-1}%
  \endgroup
}
\makeatother
\title{AraBART: a Pretrained Arabic Sequence-to-Sequence Model for Abstractive Summarization}

\author[1]{Moussa Kamal Eddine}
\author[2]{Nadi Tomeh}
\author[3]{Nizar Habash}
\author[2]{Joseph Le Roux}
\author[1,4]{Michalis Vazirgiannis}
\affil[1]{\'Ecole Polytechnique, $^\mathrm{2}$Université Sorbonne Paris Nord, $^\mathrm{3}$New York University Abu Dhabi, $^\mathrm{4}$AUEB}

\begin{document}
\maketitle

\begin{abstract}
Like most natural language understanding and generation tasks, state-of-the-art models for summarization are transformer-based sequence-to-sequence architectures that are pretrained on large corpora.
While most existing models focused on English, Arabic remained understudied.
In this paper we propose AraBART, the first Arabic model in which the encoder and the decoder are pretrained end-to-end, based on  BART \cite{lewis-etal-2020-bart}.
We show that AraBART achieves the best performance on multiple abstractive summarization datasets, outperforming strong baselines including
a pretrained Arabic BERT-based model and multilingual mBART and mT5 models. AraBART is available at Huggingface model hub\footnote{https://huggingface.co/moussaKam/AraBART}.
\end{list}
\end{abstract}


\section{Introduction}
Summarization is the task of transforming a text into a shorter 
representation of its essential meaning in natural language.
Extractive approaches \cite{nallapati-2017-summarunner,narayan-etal-2018-ranking,zhou-etal-2018-neural-document,see-etal-2017-get} identify informative spans in the original text and stitch them together to generate the summary. Abstractive approaches on the other hand are not restricted to the input \cite{rush-etal-2015-neural,chopra-etal-2016-abstractive,dou-etal-2021-gsum}.

While the vast majority of published models in both categories focused on English, some tackled other languages including Chinese \cite{hu-etal-2015-lcsts} and French \cite{kamal-eddine-etal-2021-barthez}, while Arabic remained understudied.
In fact, most Arabic summarization models are extractive \cite{qassem-etal-2019-automatic,alshanqiti2021leveraging}. They generate explainable and factual summaries but tend to be verbose and lack fluency.
Addressing this problem, abstractive models are flexible in their word choices, resorting to paraphrasing and generalization to obtain more fluent and coherent summaries. Sequence-to-sequence (seq2seq) is the architecture of choice for abstractive models.
\citet{AlMaleh2020ArabicTS}, for instance, apply the pointer-generator network \cite{see-etal-2017-get} to Arabic, while \citet{wazery-2022-text} propose a more generic RNN-based model.

There are, however, two main issues with abstractive models as applied to Arabic.
First, they are trained and evaluated either on extractive datasets such as KALIMAT \cite{el-haj-13-kalimat} and ANT Corpus \cite{ant-corpus-21}, or on headline generation datasets such as AHS \cite{AlMaleh2020ArabicTS}, which only contains short and rather extractive headlines.
Second, despite their state-of-the-art performance, abstractive models frequently generate content that is nonfactual or unfaithful to the original text.
\citet{maynez-etal-2020-faithfulness} showed that English models that are based on the Transformer architecture such as \textsc{Bert2Bert} \cite{rothe-etal-2020-leveraging}
efficiently mitigate this phenomenon thanks to pretraining on large corpora.
Therefore, \citet{elmadani2020bert} finetuned a pretrained BERT
using the encoder-decoder architecture of \textsc{BertSum} \cite{liu-lapata-2019-text}. However, only the encoder is pretrained, the decoder and the connection weights between the encoder and the decoder are initialized randomly which is sub-optimal.

To address these two problems, we propose AraBART, the first sequence-to-sequence Arabic model in which the encoder, the decoder and their connection weights are pretrained end-to-end using BART's denoising autoencoder objective \cite{lewis-etal-2020-bart}.
While the encoder is bidirectional, the decoder is auto-regressive and thus more suitable for summarization than BERT-based decoders.
We finetuned and evaluate our model on two abstractive datasets. The first is Arabic Gigaword \cite{ar-gigaword}, a newswire headline-generation dataset, not previously exploited in Arabic abstractive summarization; the second is XL-Sum, a multilingual text summarization dataset for 44 languages including Arabic \cite{hasan-etal-2021-xl}. AraBART achieves state-of-the-art results outperforming pretrained BERT-based models as well as a much larger model, mBART25 \cite{liu-etal-2020-multilingual-denoising}, a multilingual denoising auto-encoder pretrained on 25 different languages using the BART objective.

In section \ref{sec:AraBART} we present the architecture and the pretraining settings of AraBART. In section \ref{sec:experiments} we evaluate and compare AraBART against three strong baselines on a wide range of abstractive summarization datasets.
Finally, we discuss related work in section~\ref{related}.

\begin{table*}[]
\centering
\resizebox{\textwidth}{!}{%
\begin{tabular}{lccccccccccc}
 &
  \multicolumn{1}{l}{} &
  \multicolumn{10}{c}{\textbf{Datasets}} \\ \cline{3-12} 
\textbf{} &
  \multicolumn{1}{c|}{\textbf{}} &
  \multicolumn{1}{c|}{\textit{\textbf{AAW}}} &
  \multicolumn{1}{c|}{\textit{\textbf{AHR}}} &
  \multicolumn{1}{c|}{\textit{\textbf{AFP}}} &
  \multicolumn{1}{c|}{\textit{\textbf{HYT}}} &
  \multicolumn{1}{c|}{\textit{\textbf{NHR}}} &
  \multicolumn{1}{c|}{\textit{\textbf{QDS}}} &
  \multicolumn{1}{c|}{\textit{\textbf{XIN}}} &
  \multicolumn{1}{c|}{\textit{\textbf{MIX}}} &
  \multicolumn{1}{c|}{\textit{\textbf{XL-S}}} &
  \multicolumn{1}{c|}{\textit{\textbf{XL-T}}} \\ \hline
\multicolumn{1}{|l|}{\multirow{2}{*}{\textbf{\begin{tabular}[c]{@{}l@{}}Average\\ \#tokens\end{tabular}}}} &
  \multicolumn{1}{c|}{\textit{document}} &
  \multicolumn{1}{c|}{453.3} &
  \multicolumn{1}{c|}{394.2} &
  \multicolumn{1}{c|}{232.8} &
  \multicolumn{1}{c|}{474.0} &
  \multicolumn{1}{c|}{455.9} &
  \multicolumn{1}{c|}{450.6} &
  \multicolumn{1}{c|}{187.2} &
  \multicolumn{1}{c|}{364.5} &
  \multicolumn{1}{c|}{428.7} &
  \multicolumn{1}{c|}{428.7} \\
\multicolumn{1}{|l|}{} &
  \multicolumn{1}{c|}{\textit{summary}} &
  \multicolumn{1}{c|}{15.5} &
  \multicolumn{1}{c|}{9.2} &
  \multicolumn{1}{c|}{8.3} &
  \multicolumn{1}{c|}{11.2} &
  \multicolumn{1}{c|}{10.4} &
  \multicolumn{1}{c|}{8.0} &
  \multicolumn{1}{c|}{8.2} &
  \multicolumn{1}{c|}{9.4} &
  \multicolumn{1}{c|}{25.6} &
  \multicolumn{1}{c|}{9.4} \\ \hline
\multicolumn{1}{|l|}{\multirow{3}{*}{\textbf{\begin{tabular}[c]{@{}l@{}}\%novel\\ n-grams\\ in summary\end{tabular}}}} &
  \multicolumn{1}{c|}{\textit{unigrams}} &
  \multicolumn{1}{c|}{44.2} &
  \multicolumn{1}{c|}{46.5} &
  \multicolumn{1}{c|}{30.7} &
  \multicolumn{1}{c|}{42.4} &
  \multicolumn{1}{c|}{46.5} &
  \multicolumn{1}{c|}{24.9} &
  \multicolumn{1}{c|}{26.4} &
  \multicolumn{1}{c|}{40.0} &
  \multicolumn{1}{c|}{53.5} &
  \multicolumn{1}{c|}{44.3} \\
\multicolumn{1}{|l|}{} &
  \multicolumn{1}{c|}{\textit{bigrams}} &
  \multicolumn{1}{c|}{78.5} &
  \multicolumn{1}{c|}{78.4} &
  \multicolumn{1}{c|}{63.6} &
  \multicolumn{1}{c|}{78.6} &
  \multicolumn{1}{c|}{80.7} &
  \multicolumn{1}{c|}{46.9} &
  \multicolumn{1}{c|}{48.5} &
  \multicolumn{1}{c|}{72.2} &
  \multicolumn{1}{c|}{85.8} &
  \multicolumn{1}{c|}{81.2} \\
\multicolumn{1}{|l|}{} &
  \multicolumn{1}{c|}{\textit{trigrams}} &
  \multicolumn{1}{c|}{91.2} &
  \multicolumn{1}{c|}{91.3} &
  \multicolumn{1}{c|}{81.9} &
  \multicolumn{1}{c|}{92.0} &
  \multicolumn{1}{c|}{92.8} &
  \multicolumn{1}{c|}{57.5} &
  \multicolumn{1}{c|}{60.8} &
  \multicolumn{1}{c|}{86.3} &
  \multicolumn{1}{c|}{95.2} &
  \multicolumn{1}{c|}{94.1} \\ \hline
\end{tabular}
}
\caption{Statistics of Gigaword subsets an XL-Sum summaries (XL-S) and titles (XL-T). The first two lines show the average document and summary lengths. The percentage of n-grams in the summary
that do not occur in the input article is used as a
measure of abstractiveness \cite{narayan-etal-2018-dont}.}
\label{tab:abstractivness}
\end{table*}

\section{AraBART}
\label{sec:AraBART}
AraBART follows the architecture of BART Base \cite{lewis-etal-2020-bart}, which has 6 encoder and 6 decoder layers and 768 hidden dimensions. In total AraBART has 139M parameters. We add one additional layer-normalization layer on top of the encoder and the decoder to stabilize training at FP16 precision, following \cite{liu-etal-2020-multilingual-denoising}.
We use sentencepiece \cite{kudo-richardson-2018-sentencepiece} to create the vocabulary of AraBART. We train the sentencepiece model on a randomly sampled subset of the pretraining corpus (without any preprocessing) with size 20GB. We fix the vocabulary size to 50K tokens and the character coverage to 99.99\% to avoid a high rate of unknown tokens.

\subsection{Pretraining}
We adopt the same corpus used to pretrain AraBERT \cite{antoun-etal-2020-arabert}. While \citet{antoun-etal-2020-arabert} use a preprocessed version of the corpus, we opted to reverse the preprocessing  by using a script that removes added spaces around non alphabetical characters, and also undo some words segmentation. The use of a corpus with no preprocessing, makes the text generation more natural. The size of the pretraining corpus before/after sentencepiece tokenization is 73/96 GB.
\paragraph{Pretraining Objective}
AraBART is a denoising autoencoder i.e. it learns to reconstruct a corrupted text. The noise function applied to the input text are the same as in \citet{lewis-etal-2020-bart}. The first noise function is \textit{text infilling}, where 30\% of the text is masked by replacing a number of text spans with a [MASK] token. The length of the spans is sampled from a Poisson distribution with $\lambda=3.5$. The second noise function is \textit{sentence permutation}, where the sentences of the input text are shuffled based on the full stops.

\paragraph{Pretraining Settings}
AraBART pretraining took approximately 60h. The pretraining was carried out on 128 Nvidia V100 GPUs which allowed for 25 full passes over the pretraining corpus. We used the Adam optimizer with $\epsilon = 10^{-6}$, $\beta_1=0.9$, and $\beta_2=0.98$ following \citet{liu2019roberta}. We use a warm up for 6\% of the pretraining were the learning rate linearly increases from 0 to 0.0006, then decreases linearly to reach 0 at the end of the pretraining. We fixed the update frequency to 2 and we use a dropout 0.1 in the first 20 epochs and we changed it to 0 in the last 5 epochs. Finally we used FP16 to speed-up the pretraining. The pretraining is done using Fairseq \cite{ott-etal-2019-fairseq}.

\section{Experiments}\label{sec:experiments}

\subsection{Datasets}
\label{sec:gigaword}
To evaluate our model, we use several datasets that consist mostly of news articles annotated with summaries with different level of abstractivness. The first 7 datasets (\textit{AAW}, \textit{AFP}, \textit{AHR}, \textit{HYT}, \textit{NHR}, \textit{QDS} and \textit{XIN}) are subsets of the Arabic Gigaword \cite{ar-gigaword} corpus. Each one is a different news source, composed of document-headline pairs. In all these datasets we use a train set of 50K examples, a validation set of size 5K examples and a test set of size 5K examples, selected randomly.
The \textit{MIX} dataset consists of 60K examples uniformly sampled from the union of the 7 different sources.

In addition the Arabic Gigaword corpus, we use XL-Sum \cite{hasan-etal-2021-xl}. The news articles in XL-sum are annotated with summaries and titles, thus creating two tasks: summary and title generation.

Table \ref{tab:abstractivness} shows that the different datasets used in our experiments cover a wide range of article/summary lengths and levels of abstractivness.

\subsection{Baselines}
We compare our model to three types of state-of-the-art baselines. The first, called C2C, is a monolingual seq2seq model based on \textsc{Bert2Bert} \cite{rothe-etal-2020-leveraging}. The encoder and decoder are initialized using \textsc{CamelBert} \cite{inoue-etal-2021-interplay} weights while the cross-attention weights are randomly initialized.\footnote{We experimented with \textsc{AraBERT} \cite{antoun-etal-2020-arabert} which was slower to converge and didn't achieve better performance.} C2C has 246M parameters in total.  

The second baseline is mBART25 \cite{liu-etal-2020-multilingual-denoising} which is a multilingual BART pretrained on 25 different languages including Arabic. Although mBART25 was initially pretrained for neural machine translation, it was shown that it can be used in monolingual generative tasks such as abstractive summarization \cite{kamal-eddine-etal-2021-barthez}. mBART25 has 610M parameters in total. 

While mBART25 is pretrained on multilingual corpora, we finetuned it on Arabic data only. We therefore, include a third multilingual baseline pretrained and finetuned on multilingual data. We use the checkpoint\footnote{\url{https://huggingface.co/csebuetnlp/mT5_multilingual_XLSum}} of mT5$_{base}$ in the comparison on XL-S (summary). This checkpoint was finetuned on the training set of the 45 different languages included in the corpus. The total training size is 1M multilingual examples shuffled together \cite{hasan-etal-2021-xl}. mT5$_{base}$ has 582M parameters in total.

\subsection{Training and Evaluation}
We finetuned each model for three epochs, using the Adam optimizer and $5\times10^{-5}$ maximum learning rate with linear decay scheduling. In the generation phase we use beam-search with beam size of~3.

For evaluation, we first normalize the output summaries as is standard practice in Arabic: we removed Tatweel and diacritization, we normalized Alef/Yaa and separated punctuations.
We report ROUGE-1, ROUGE-2 and ROUGE-L f1-scores \cite{lin-2004-rouge}.
However, these metrics are solely based on surface-form matching and have limited sense of semantic similarity \cite{eddine2021frugalscore}. Thus we opted for using BERTScore \cite{Zhang2020BERTScore}, a metric based on the similarity of the contextual embeddings of the reference and candidate summaries, produced by a BERT-like model.\footnote{We use the official implementation (\url{https://github.com/Tiiiger/bert\_score}) with the following options: \texttt{-m UBC-NLP/ARBERT -l 9} \cite{chiang-etal-2020-pretrained}}

\hide{
 
\begin{table}[!ht]\centering
\def\arraystretch{1.4}
\footnotesize
\begin{tabular}{|llrrrr|}
\hline
\textbf{Dataset} & \textbf{Model} & \textbf{R1} & \textbf{R2} & \textbf{RL} & \textbf{BS} \\
\hline
\multirow{3}{*}{aaw} & AraBART & 30.7&	15.3	&27.4	&62.5 \\
 & mBART25 & 29.5 & 	14.35& 26.0&	61.5 \\
 & C2C & 24.6 &	9.87 & 21.7 &	58.3 \\
 \hline
\multirow{3}{*}{afp} & AraBART & 55.0 & 37.9 & 53.4 & 77.5 \\
 & mBART25 & 54.8 & 37.3 & 52.8 & 77.2 \\
 & C2C & 50.0 & 32.2 & 48.4 & 74.8 \\
\hline
\multirow{3}{*}{ahr} & AraBART & 39.1 & 25.4 & 37.7 & 68.2 \\
 & mBART25 & 39.1 & 26.1 & 37.5 & 68.1 \\
 & C2C & 33.0 & 19.7 & 31.8 & 63.5 \\
\hline
\multirow{3}{*}{hyt} & AraBART & 33.1 & 17.5 & 30.7 & 63.8 \\
 & mBART25 & 32.0 & 16.2 & 29.3 & 63.1 \\
 & C2C & 27.4 & 11.5 & 25.2 & 59.6 \\
\hline
\multirow{3}{*}{nhr} & AraBART & 32.0 & 17.2 & 30.3 & 61.2 \\
 & mBART25 & 31.0 & 16.2 & 29.2 & 60.3 \\
 & C2C & 24.1 & 10.0 & 22.9 & 53.0 \\
\hline
\multirow{3}{*}{qds} & AraBART & 62.1 & 53.9 & 61.4 & 80.3 \\
 & mBART25 & 62.4 & 54.1 & 61.7 & 80.4 \\
 & C2C & 57.9 & 48.9 & 57.4 & 77.3 \\
\hline
\multirow{3}{*}{xin} & AraBART & 66.0 & 53.9 & 65.1 & 84.4 \\
 & mBART25 & 65.1 & 53.4 & 64.2 & 84.0 \\
 & C2C & 62.4 & 50.1 & 61.6 & 82.5 \\
\hline
\multirow{3}{*}{mix} & AraBART & 39.2 & 25.5 & 37.6 & 67.6 \\
 & mBART25 & 39.0 & 25.6 & 37.1 & 67.2 \\
 & C2C & 32.8 & 19.1 & 31.4 & 62.5 \\
\hline
\multirow{3}{*}{\makecell{XL-Sum\\(summary)}} & AraBART & 34.5 & 14.6 & 30.5 & 67.0 \\
 & mBART25 & 32.1 & 12.5 & 27.6 & 65.3 \\
 & C2C & 26.9 & 8.7 & 23.1 & 61.6 \\
 & mT5${_base}$ & 26.9 & 8.7 & 23.1 & 61.6 \\
\hline
\multirow{3}{*}{\makecell{XL-Sum\\(title)}} & AraBART & 32.0 & 13.7 & 29.4 & 65.8 \\
 & mBART25 & 29.8 & 11.7 & 26.9 & 64.3 \\
 & C2C & 25.2 & 7.9 & 22.9 & 61.1 \\
\hline
\hline
\multirow{3}{*}{\makecell{Macro\\(average)}} & AraBART & 42.4 & 28.8 & 40.3 & 69.8 \\
 & mBART25 & 41.5 & 28.1 & 39.2 & 69.1 \\
 & C2C & 36.4 & 23.1 & 34.6 & 65.4 \\
\hline
\end{tabular}
\end{table}
}

\begin{table}[!t]
\centering
\setlength{\tabcolsep}{3.5pt}
\begin{tabular}{|c|lrrrr|}
\hline
\multicolumn{1}{|c|}{\textbf{Source}} & \multicolumn{1}{|l|}{\textbf{Model}} & \multicolumn{1}{r|}{\textbf{R1}}   & \multicolumn{1}{r|}{\textbf{R2}}   & \multicolumn{1}{r|}{\textbf{RL}}   & \textbf{BS}   \\ \hline\hline
\multicolumn{1}{|c|}{\textit{\textbf{AAW}}} &                                                                                                        
\multicolumn{1}{|l|}{AraBART}        & \multicolumn{1}{r|}{\textbf{30.7}} & \multicolumn{1}{r|}{\textbf{15.3}} & \multicolumn{1}{r|}{\textbf{27.4}} & \textbf{62.5} \\ \cline{2-6}
&\multicolumn{1}{|l|}{mBART25}          & \multicolumn{1}{r|}{29.5}          & \multicolumn{1}{r|}{14.4}          & \multicolumn{1}{r|}{26.0}          & 61.5          \\ \cline{2-6}
&\multicolumn{1}{|l|}{C2C}            & \multicolumn{1}{r|}{24.6}          & \multicolumn{1}{r|}{9.87}          & \multicolumn{1}{r|}{21.7}          & 58.3          \\ \hline\hline
\multicolumn{1}{|c|}{\textit{\textbf{AFP}}}   &                                                                                                      
\multicolumn{1}{|l|}{AraBART}        & \multicolumn{1}{r|}{\textbf{55.0}} & \multicolumn{1}{r|}{\textbf{37.9}} & \multicolumn{1}{r|}{\textbf{53.4}} & \textbf{77.5} \\ \cline{2-6}
&\multicolumn{1}{|l|}{mBART25}          & \multicolumn{1}{r|}{54.8}          & \multicolumn{1}{r|}{37.3}          & \multicolumn{1}{r|}{52.8}          & 77.2          \\ \cline{2-6}
&\multicolumn{1}{|l|}{C2C}            & \multicolumn{1}{r|}{50.0}          & \multicolumn{1}{r|}{32.2}          & \multicolumn{1}{r|}{48.4}          & 74.8          \\ \hline\hline
\multicolumn{1}{|c|}{\textit{\textbf{AHR}}}     &                                                                                                    
\multicolumn{1}{|l|}{AraBART}        & \multicolumn{1}{r|}{\textbf{39.1}} & \multicolumn{1}{r|}{25.4}          & \multicolumn{1}{r|}{\textbf{37.7}} & \textbf{68.2} \\ \cline{2-6}
&\multicolumn{1}{|l|}{mBART25}          & \multicolumn{1}{r|}{\textbf{39.1}} & \multicolumn{1}{r|}{\textbf{26.1}} & \multicolumn{1}{r|}{37.5}          & 68.1          \\ \cline{2-6}
&\multicolumn{1}{|l|}{C2C}            & \multicolumn{1}{r|}{33.0}          & \multicolumn{1}{r|}{19.7}          & \multicolumn{1}{r|}{31.8}          & 63.5          \\ \hline\hline
\multicolumn{1}{|c|}{\textit{\textbf{HYT}}}     &                                                                                                    
\multicolumn{1}{|l|}{AraBART}        & \multicolumn{1}{r|}{\textbf{33.1}} & \multicolumn{1}{r|}{\textbf{17.5}} & \multicolumn{1}{r|}{\textbf{30.7}} & \textbf{63.8} \\ \cline{2-6}
&\multicolumn{1}{|l|}{mBART25}          & \multicolumn{1}{r|}{32.0}          & \multicolumn{1}{r|}{16.2}          & \multicolumn{1}{r|}{29.3}          & 63.1          \\ \cline{2-6}
&\multicolumn{1}{|l|}{C2C}            & \multicolumn{1}{r|}{27.4}          & \multicolumn{1}{r|}{11.5}          & \multicolumn{1}{r|}{25.2}          & 59.6          \\ \hline\hline
\multicolumn{1}{|c|}{\textit{\textbf{NHR}}}    &                                                                                                     
\multicolumn{1}{|l|}{AraBART}        & \multicolumn{1}{r|}{\textbf{32.0}} & \multicolumn{1}{r|}{\textbf{17.2}} & \multicolumn{1}{r|}{\textbf{30.3}} & \textbf{61.2} \\ \cline{2-6}
&\multicolumn{1}{|l|}{mBART25}          & \multicolumn{1}{r|}{31.0}          & \multicolumn{1}{r|}{16.2}          & \multicolumn{1}{r|}{29.2}          & 60.3          \\ \cline{2-6}
&\multicolumn{1}{|l|}{C2C}            & \multicolumn{1}{r|}{24.1}          & \multicolumn{1}{r|}{10.0}          & \multicolumn{1}{r|}{22.9}          & 53.0          \\ \hline\hline
\multicolumn{1}{|c|}{\textit{\textbf{QDS}}}   &                                                                                                      
\multicolumn{1}{|l|}{AraBART}        & \multicolumn{1}{r|}{62.1}          & \multicolumn{1}{r|}{53.9}          & \multicolumn{1}{r|}{61.4}          & 80.3          \\ \cline{2-6}
&\multicolumn{1}{|l|}{mBART25}          & \multicolumn{1}{r|}{\textbf{62.4}} & \multicolumn{1}{r|}{\textbf{54.1}} & \multicolumn{1}{r|}{\textbf{61.7}} & \textbf{80.4} \\ \cline{2-6}
&\multicolumn{1}{|l|}{C2C}            & \multicolumn{1}{r|}{57.9}          & \multicolumn{1}{r|}{48.9}          & \multicolumn{1}{r|}{57.4}          & 77.3          \\ \hline\hline
\multicolumn{1}{|c|}{\textit{\textbf{XIN}}}     &                                                                                                    
\multicolumn{1}{|l|}{AraBART}        & \multicolumn{1}{r|}{\textbf{66.0}} & \multicolumn{1}{r|}{\textbf{53.9}} & \multicolumn{1}{r|}{\textbf{65.1}} & \textbf{84.4} \\ \cline{2-6}
&\multicolumn{1}{|l|}{mBART25}          & \multicolumn{1}{r|}{65.1}          & \multicolumn{1}{r|}{53.4}          & \multicolumn{1}{r|}{64.2}          & 84.0          \\ \cline{2-6}
&\multicolumn{1}{|l|}{C2C}            & \multicolumn{1}{r|}{62.4}          & \multicolumn{1}{r|}{50.1}          & \multicolumn{1}{r|}{61.6}          & 82.5          \\ \hline\hline
\multicolumn{1}{|c|}{\textit{\textbf{MIX}}}     &                                                                                                    
\multicolumn{1}{|l|}{AraBART}        & \multicolumn{1}{r|}{\textbf{39.2}} & \multicolumn{1}{r|}{25.5}          & \multicolumn{1}{r|}{\textbf{37.6}} & \textbf{67.6} \\ \cline{2-6}
&\multicolumn{1}{|l|}{mBART25}          & \multicolumn{1}{r|}{39.0}          & \multicolumn{1}{r|}{\textbf{25.6}} & \multicolumn{1}{r|}{37.1}          & 67.2          \\ \cline{2-6}
&\multicolumn{1}{|l|}{C2C}            & \multicolumn{1}{r|}{32.8}          & \multicolumn{1}{r|}{19.1}          & \multicolumn{1}{r|}{31.4}          & 62.5          \\ \hline\hline
\multicolumn{1}{|c|}{\textit{\textbf{XL-S}}}       &                                                                                   
\multicolumn{1}{|l|}{AraBART}        & \multicolumn{1}{r|}{\textbf{34.5}} & \multicolumn{1}{r|}{\textbf{14.6}} & \multicolumn{1}{r|}{\textbf{30.5}} & \textbf{67.0} \\ \cline{2-6}
&\multicolumn{1}{|l|}{mBART25}          & \multicolumn{1}{r|}{32.1}          & \multicolumn{1}{r|}{12.5}          & \multicolumn{1}{r|}{27.6}          & 65.3          \\ \cline{2-6}
&\multicolumn{1}{|l|}{C2C}            & \multicolumn{1}{r|}{26.9}          & \multicolumn{1}{r|}{8.7}           & \multicolumn{1}{r|}{23.1} 
& 61.6          \\ \cline{2-6}
&\multicolumn{1}{|l|}{mT5$_{base}$}            & \multicolumn{1}{r|}{32.8}          & \multicolumn{1}{r|}{12.7}           & \multicolumn{1}{r|}{28.7} 
& 65.8          \\ 
   
\hline\hline
\multicolumn{1}{|c|}{\textit{\textbf{XL-T}}}      &                                                                                       
\multicolumn{1}{|l|}{AraBART}        & \multicolumn{1}{r|}{\textbf{32.0}} & \multicolumn{1}{r|}{\textbf{13.7}} & \multicolumn{1}{r|}{\textbf{29.4}} & \textbf{65.8} \\ \cline{2-6}
 &\multicolumn{1}{|l|}{mBART25}          & \multicolumn{1}{r|}{29.8}          & \multicolumn{1}{r|}{11.7}          & \multicolumn{1}{r|}{26.9}          & 64.3          \\ \cline{2-6}
&\multicolumn{1}{|l|}{C2C}            & \multicolumn{1}{r|}{25.2}          & \multicolumn{1}{r|}{7.9}           & \multicolumn{1}{r|}{22.9}          & 61.1          \\ \hline\hline
\multicolumn{1}{|c|}{\textit{\textbf{Macro}}}    &                                                                                           \multicolumn{1}{|l|}{AraBART}        & \multicolumn{1}{r|}{\textbf{42.4}} & \multicolumn{1}{r|}{\textbf{28.8}} & \multicolumn{1}{r|}{\textbf{40.3}} & \textbf{69.8} \\ \cline{2-6}
\multicolumn{1}{|c|}{\textit{\textbf{Averages}}}  &\multicolumn{1}{|l|}{mBART25}          & \multicolumn{1}{r|}{41.5}          & \multicolumn{1}{r|}{28.1}          & \multicolumn{1}{r|}{39.2}          & 69.1          \\ \cline{2-6}
&\multicolumn{1}{|l|}{C2C}            & \multicolumn{1}{r|}{36.4}          & \multicolumn{1}{r|}{23.1}          & \multicolumn{1}{r|}{34.6}          & 65.4          \\ \hline
\end{tabular}
\caption{The performance of AraBART, mBART25 and C2C (CamelBert2CamelBert) on all datasets in terms of ROUGE-1 (R1), ROUGE-2 (R2), ROUGE-L (RL) and BERTScore (BS). Macro averages are computed over all datasets.}
\label{tab:main-results}
\end{table}

\hide{
\begin{table}[]
\centering
\begin{tabular}{|lrrrr|}
\hline
\multicolumn{1}{|l|}{\textbf{Model}} & \multicolumn{1}{r|}{\textbf{R1}}   & \multicolumn{1}{r|}{\textbf{R2}}   & \multicolumn{1}{r|}{\textbf{RL}}   & \textbf{BS}   \\ \hline
\multicolumn{5}{|c|}{\textit{AAW}}                                                                                                                                  \\ \hline
\multicolumn{1}{|l|}{AraBART}        & \multicolumn{1}{r|}{\textbf{30.7}} & \multicolumn{1}{r|}{\textbf{15.3}} & \multicolumn{1}{r|}{\textbf{27.4}} & \textbf{62.5} \\ \hline
\multicolumn{1}{|l|}{mBART25}          & \multicolumn{1}{r|}{29.5}          & \multicolumn{1}{r|}{14.4}          & \multicolumn{1}{r|}{26.0}          & 61.5          \\ \hline
\multicolumn{1}{|l|}{C2C}            & \multicolumn{1}{r|}{24.6}          & \multicolumn{1}{r|}{9.87}          & \multicolumn{1}{r|}{21.7}          & 58.3          \\ \hline
\multicolumn{5}{|c|}{\textit{AFP}}                                                                                                                                  \\ \hline
\multicolumn{1}{|l|}{AraBART}        & \multicolumn{1}{r|}{\textbf{55.0}} & \multicolumn{1}{r|}{\textbf{37.9}} & \multicolumn{1}{r|}{\textbf{53.4}} & \textbf{77.5} \\ \hline
\multicolumn{1}{|l|}{mBART25}          & \multicolumn{1}{r|}{54.8}          & \multicolumn{1}{r|}{37.3}          & \multicolumn{1}{r|}{52.8}          & 77.2          \\ \hline
\multicolumn{1}{|l|}{C2C}            & \multicolumn{1}{r|}{50.0}          & \multicolumn{1}{r|}{32.2}          & \multicolumn{1}{r|}{48.4}          & 74.8          \\ \hline
\multicolumn{5}{|c|}{\textit{AHR}}                                                                                                                                  \\ \hline
\multicolumn{1}{|l|}{AraBART}        & \multicolumn{1}{r|}{\textbf{39.1}} & \multicolumn{1}{r|}{25.4}          & \multicolumn{1}{r|}{\textbf{37.7}} & \textbf{68.2} \\ \hline
\multicolumn{1}{|l|}{mBART25}          & \multicolumn{1}{r|}{\textbf{39.1}} & \multicolumn{1}{r|}{\textbf{26.1}} & \multicolumn{1}{r|}{37.5}          & 68.1          \\ \hline
\multicolumn{1}{|l|}{C2C}            & \multicolumn{1}{r|}{33.0}          & \multicolumn{1}{r|}{19.7}          & \multicolumn{1}{r|}{31.8}          & 63.5          \\ \hline
\multicolumn{5}{|c|}{\textit{HYT}}                                                                                                                                  \\ \hline
\multicolumn{1}{|l|}{AraBART}        & \multicolumn{1}{r|}{\textbf{33.1}} & \multicolumn{1}{r|}{\textbf{17.5}} & \multicolumn{1}{r|}{\textbf{30.7}} & \textbf{63.8} \\ \hline
\multicolumn{1}{|l|}{mBART25}          & \multicolumn{1}{r|}{32.0}          & \multicolumn{1}{r|}{16.2}          & \multicolumn{1}{r|}{29.3}          & 63.1          \\ \hline
\multicolumn{1}{|l|}{C2C}            & \multicolumn{1}{r|}{27.4}          & \multicolumn{1}{r|}{11.5}          & \multicolumn{1}{r|}{25.2}          & 59.6          \\ \hline
\multicolumn{5}{|c|}{\textit{NHR}}                                                                                                                                  \\ \hline
\multicolumn{1}{|l|}{AraBART}        & \multicolumn{1}{r|}{\textbf{32.0}} & \multicolumn{1}{r|}{\textbf{17.2}} & \multicolumn{1}{r|}{\textbf{30.3}} & \textbf{61.2} \\ \hline
\multicolumn{1}{|l|}{mBART25}          & \multicolumn{1}{r|}{31.0}          & \multicolumn{1}{r|}{16.2}          & \multicolumn{1}{r|}{29.2}          & 60.3          \\ \hline
\multicolumn{1}{|l|}{C2C}            & \multicolumn{1}{r|}{24.1}          & \multicolumn{1}{r|}{10.0}          & \multicolumn{1}{r|}{22.9}          & 53.0          \\ \hline
\multicolumn{5}{|c|}{\textit{QDS}}                                                                                                                                  \\ \hline
\multicolumn{1}{|l|}{AraBART}        & \multicolumn{1}{r|}{62.1}          & \multicolumn{1}{r|}{53.9}          & \multicolumn{1}{r|}{61.4}          & 80.3          \\ \hline
\multicolumn{1}{|l|}{mBART25}          & \multicolumn{1}{r|}{\textbf{62.4}} & \multicolumn{1}{r|}{\textbf{54.1}} & \multicolumn{1}{r|}{\textbf{61.7}} & \textbf{80.4} \\ \hline
\multicolumn{1}{|l|}{C2C}            & \multicolumn{1}{r|}{57.9}          & \multicolumn{1}{r|}{48.9}          & \multicolumn{1}{r|}{57.4}          & 77.3          \\ \hline
\multicolumn{5}{|c|}{\textit{XIN}}                                                                                                                                  \\ \hline
\multicolumn{1}{|l|}{AraBART}        & \multicolumn{1}{r|}{\textbf{66.0}} & \multicolumn{1}{r|}{\textbf{53.9}} & \multicolumn{1}{r|}{\textbf{65.1}} & \textbf{84.4} \\ \hline
\multicolumn{1}{|l|}{mBART25}          & \multicolumn{1}{r|}{65.1}          & \multicolumn{1}{r|}{53.4}          & \multicolumn{1}{r|}{64.2}          & 84.0          \\ \hline
\multicolumn{1}{|l|}{C2C}            & \multicolumn{1}{r|}{62.4}          & \multicolumn{1}{r|}{50.1}          & \multicolumn{1}{r|}{61.6}          & 82.5          \\ \hline
\multicolumn{5}{|c|}{\textit{MIX}}                                                                                                                                  \\ \hline
\multicolumn{1}{|l|}{AraBART}        & \multicolumn{1}{r|}{\textbf{39.2}} & \multicolumn{1}{r|}{25.5}          & \multicolumn{1}{r|}{\textbf{37.6}} & \textbf{67.6} \\ \hline
\multicolumn{1}{|l|}{mBART25}          & \multicolumn{1}{r|}{39.0}          & \multicolumn{1}{r|}{\textbf{25.6}} & \multicolumn{1}{r|}{37.1}          & 67.2          \\ \hline
\multicolumn{1}{|l|}{C2C}            & \multicolumn{1}{r|}{32.8}          & \multicolumn{1}{r|}{19.1}          & \multicolumn{1}{r|}{31.4}          & 62.5          \\ \hline
\multicolumn{5}{|c|}{\textit{XL-Sum (summaries)}}                                                                                                                   \\ \hline
\multicolumn{1}{|l|}{AraBART}        & \multicolumn{1}{r|}{\textbf{34.5}} & \multicolumn{1}{r|}{\textbf{14.6}} & \multicolumn{1}{r|}{\textbf{30.5}} & \textbf{67.0} \\ \hline
\multicolumn{1}{|l|}{mBART25}          & \multicolumn{1}{r|}{32.1}          & \multicolumn{1}{r|}{12.5}          & \multicolumn{1}{r|}{27.6}          & 65.3          \\ \hline
\multicolumn{1}{|l|}{C2C}            & \multicolumn{1}{r|}{26.9}          & \multicolumn{1}{r|}{8.7}           & \multicolumn{1}{r|}{23.1}          & 61.6          \\ \hline
\multicolumn{5}{|c|}{\textit{XL-Sum (titles)}}                                                                                                                      \\ \hline
\multicolumn{1}{|l|}{AraBART}        & \multicolumn{1}{r|}{\textbf{32.0}} & \multicolumn{1}{r|}{\textbf{13.7}} & \multicolumn{1}{r|}{\textbf{29.4}} & \textbf{65.8} \\ \hline
\multicolumn{1}{|l|}{mBART25}          & \multicolumn{1}{r|}{29.8}          & \multicolumn{1}{r|}{11.7}          & \multicolumn{1}{r|}{26.9}          & 64.3          \\ \hline
\multicolumn{1}{|l|}{C2C}            & \multicolumn{1}{r|}{25.2}          & \multicolumn{1}{r|}{7.9}           & \multicolumn{1}{r|}{22.9}          & 61.1          \\ \hline
\multicolumn{5}{|c|}{\textit{Macro Averages}}                                                                                                                       \\ \hline
\multicolumn{1}{|l|}{AraBART}        & \multicolumn{1}{r|}{\textbf{42.4}} & \multicolumn{1}{r|}{\textbf{28.8}} & \multicolumn{1}{r|}{\textbf{40.3}} & \textbf{69.8} \\ \hline
\multicolumn{1}{|l|}{mBART25}          & \multicolumn{1}{r|}{41.5}          & \multicolumn{1}{r|}{28.1}          & \multicolumn{1}{r|}{39.2}          & 69.1          \\ \hline
\multicolumn{1}{|l|}{C2C}            & \multicolumn{1}{r|}{36.4}          & \multicolumn{1}{r|}{23.1}          & \multicolumn{1}{r|}{34.6}          & 65.4          \\ \hline
\end{tabular}
\caption{The performance of AraBART, mBART25 and C2C (CamelBert2CamelBert) on all datasets in terms of ROUGE-1 (R1), ROUGE-2 (R2), ROUGE-L (RL) and BERTScore (BS). Macro averages are computed over all datasets.}
\label{tab:main-results}
\end{table}

}

\subsection{Results}
We observe in Table \ref{tab:main-results} that AraBART outperforms C2C on all datasets with a clear margin. This is probably a direct consequence of pretraining the seq2seq architecture end-to-end.

AraBART also outperforms mBART25 on XL-Sum which is the most abstractive dataset.
On Gigawords, AraBART is best everywhere except on AHR with mitigated results. On QDS, however, it falls clearly behind mBART25 on all metrics.
In fact, we notice that the gap between AraBART and the baselines is greater on the XL-Sum dataset than Gigaword. For instance, our model's ROUGE-L score is 2.9 absolute points higher that mBART25 on XL-S while the maximum margin obtained on a Gigaword subset is 1.4 points on AAW and HYT. We observe a tendency for AraBART to outperform mBART on more abstractive datasets. In fact, the margin between their BERTScores is positively correlated with abstractiveness as measures by the percentage of novel trigrams.\footnote{With a Pearson R score of 0.6625 and $p$-value<0.05.}

On the XL-Sum dataset, AraBART also outperforms mT5 which was finetuned in multilingual setup using more data \cite{hasan-etal-2021-xl}.

Figure~\ref{fig:example} presents some examples of the output of the various systems we studied.


\section{Related Work}
\label{related}
\paragraph{Arabic Summarization}
The overwhelming majority of past Arabic models are extractive \cite{douzidia04lakhas,ikhtisar09,ElHaj2011MultidocumentAT,shishtawy12key,haboush-12-cluster,belkebir-15-adaboost,Qaroush2021AnES,ayed-21-knapsack}.
Recently, seq2seq abstractive models for Arabic have been proposed in the literature \cite{AlMaleh2020ArabicTS,suleiman20deep,wazery-2022-text}, but none of them used pretraining.
Fine-tuning Transformer-based language models like BERT \cite{devlin-etal-2019-bert} has been shown to help Arabic abstractive \cite{elmadani2020bert} and extractive \cite{helmy-18-applying} summarization, but unlike AraBART, not all components of the model are pretrained.
Readily-available multilingual pretrained seq2seq models have been applied to Arabic summarization. \citet{kahla-etal-2021-cross} uses mBART25 \cite{liu-etal-2020-multilingual-denoising} in cross-lingual transfer setup on an unpublished dataset, while \citet{hasan-etal-2021-xl} experiment with mT5 \cite{xue-etal-2021-mt5} on XL-Sum.
Our model, tailored specifically for Arabic, outperform mBART25 and mT5 for almost all datasets despite having a smaller architecture with less parameters.

\paragraph{Arabic Datasets}
Most available datasets for Arabic are extractive \cite{el-haj-10-essex,ant-corpus-21}, use short headlines that are designed to attract the reader \cite{webz, AlMaleh2020ArabicTS}, or contain machine-generated \cite{el-haj-13-kalimat} or translated \cite{el-haj-12-english} summaries.
Notable exceptions we choose for our experiments are Gigaword \cite{ar-gigaword} and XL-Sum \cite{hasan-etal-2021-xl} because they cover both headline and summary generation, contains multiple sources, and manifest variable levels of abstractivness as shown in Table \ref{tab:abstractivness}.

\paragraph{Pretrained seq2seq models} BART-based models have been developed for multiple language including English \cite{lewis-etal-2020-bart}, French \cite{kamal-eddine-etal-2021-barthez} and Chinese \cite{shao2021cpt} in addition to multilingual models \cite{liu-etal-2020-multilingual-denoising}. While they can be finetuned to perform any language understanding or generation tasks, we focus on summarization in this work.

\section{Conclusion and Future Work}
We release AraBART, the first sequenece-to-sequence pretrained Arabic model. We evaluated our model on a set of abstractive summarization tasks, with different level of abstractiveness. We compared AraBART to two state-of-the-art models and we showed that it outperforms them almost everywhere despite the fact that it is smaller in terms of parameters.
In future work, we are planning to extend the model to multitask setups to take advantage of availability of both titles and summaries in some datasets including XL-Sum, and use external knowledge sources to improve faithfulness.
We will also explore new directions for evaluating summarization on morphologically rich languages like Arabic.

\section*{Ethical Considerations}

\paragraph{Limitations}
Our models are optimized for news text summarization; we do not expect compararble performance on other summarization tasks without additional training data.
\paragraph{Risks}
We acknowledge that our models sometimes produce incorrect non-factual and non-grammatical output, which can be misleading to general users.
\paragraph{Data}
All of the data we used comes from reputable news agencies and do not contain unanonymized private information or malicious social media content. 
\paragraph{Models}
We will make our pretrained and finetuned models available on the well known Hugging Face models hub\footnote{\url{https://huggingface.co/models}}, so it can be easily used and distributed for research or production purposes.

\bibliography{anthology,custom}
\bibliographystyle{acl_natbib}

\appendix
\section{Example Appendix}
\label{sec:appendix}
\begin{figure*}[t!]
\centering
\includegraphics[width=0.9\textwidth]{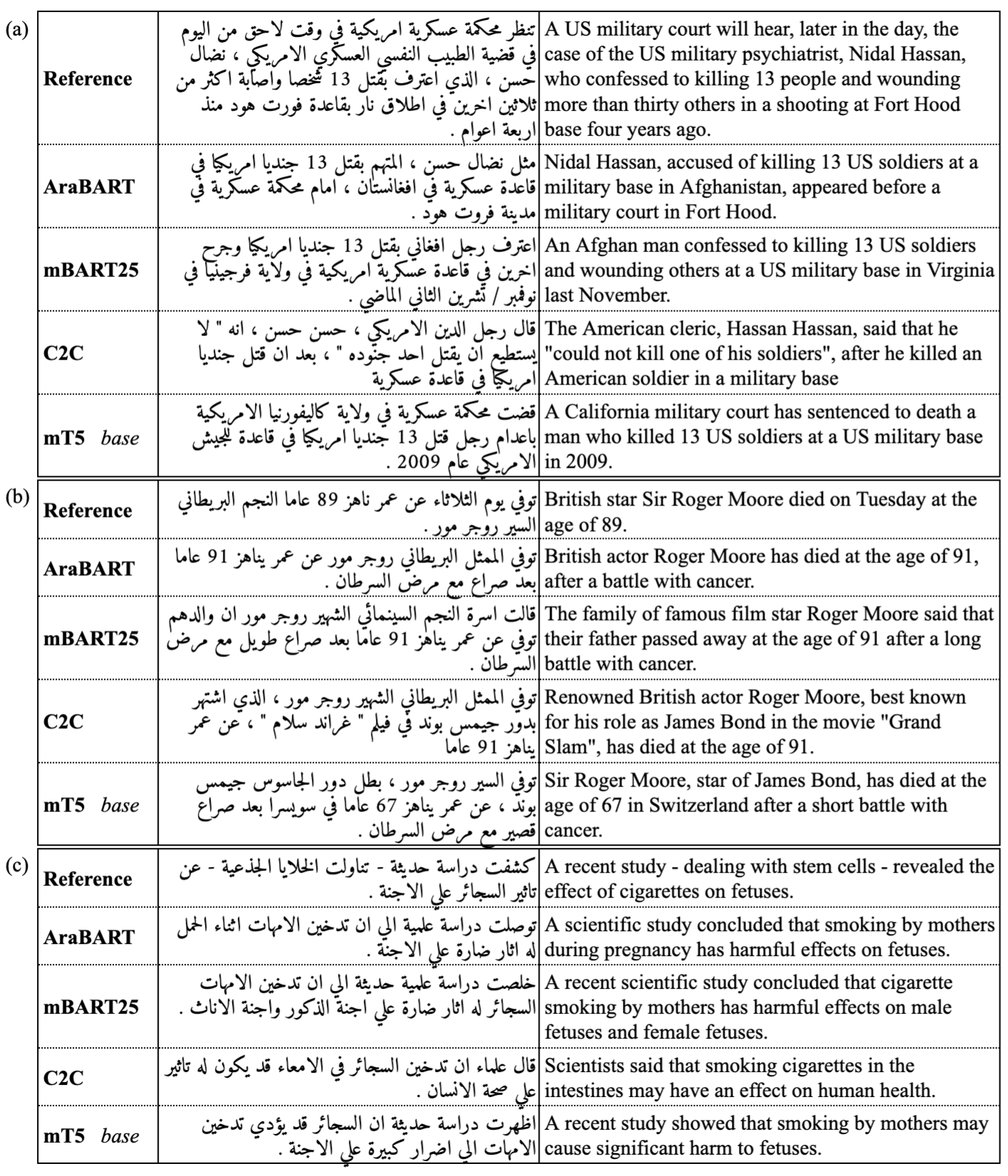}
\caption{Three selected examples contrasting the output of the various systems we studied. All examples are from the XL-Sum summaries test set.  We provide English translations to provide context for the general readers. }
\label{fig:example}
\end{figure*}

Figure~\ref{fig:example} presents some examples of the output of the various systems we studied.


\end{document}